\documentclass[sigconf]{acmart}

\usepackage{colortbl}  
\usepackage{tcolorbox}
\usepackage{balance}
\usepackage{xcolor}
\usepackage{array}  
\usepackage{graphicx}
\usepackage{array} % For better column definitions, though not strictly needed for this
\usepackage{makecell} % For multi-line cells like the first entry
\usepackage{booktabs}
\usepackage{multirow}
\usepackage{enumitem}
\AtBeginDocument{%
  }

\copyrightyear{2025} 
\acmYear{2025} 
\setcopyright{acmlicensed}
\acmConference[MM '25]{Proceedings of the 33rd ACM International Conference on Multimedia}{October 27--31, 2025}{Dublin, Ireland}
\acmBooktitle{Proceedings of the 33rd ACM International Conference on Multimedia (MM '25), October 27--31, 2025, Dublin, Ireland}
\acmDOI{10.1145/3746027.3758212}
\acmISBN{979-8-4007-2035-2/2025/10}

\begin{document}

\title{SpineBench: Benchmarking Multimodal LLMs for Spinal Pathology Analysis}

\author{Chenghanyu Zhang}
\authornote{Both authors contributed equally to this work.}
\email{zhangchenghanyu@bupt.edu.cn}
% \orcid{0009-0000-8541-4488}
\affiliation{%
  \institution{Beijing University of Posts and Telecommunications}
   \city{Beijing}
  \state{}
  \country{China}
}

\author{Zekun Li}
\authornotemark[1]
\email{zekunli@cs.ucsb.edu}
\affiliation{%
  \institution{University of California, Santa Barbara}
   \city{Santa Barbara}
  \state{California}
  \country{United States}
  }

\author{Peipei Li}
\authornote{Corresponding authors to this work.}
\email{lipeipei@bupt.edu.cn}
\affiliation{%
  \institution{Beijing University of Posts and Telecommunications}
   \city{Beijing}
  \state{}
  \country{China}
  }

\author{Xing Cui}
\email{cuixing@bupt.edu.cn}
\affiliation{%
  \institution{Beijing University of Posts and Telecommunications}
   \city{Beijing}
  \state{}
  \country{China}
  }

\author{Shuhan Xia}
\email{shuhanxia@bupt.edu.cn}
\affiliation{%
  \institution{Beijing University of Posts and Telecommunications}
   \city{Beijing}
  \state{}
  \country{China}
  }

\author{Weixiang Yan}
\email{weixiangyan@cs.ucsb.edu}
\affiliation{%
  \institution{University of California, Santa Barbara}
   \city{Santa Barbara}
  \state{California}
  \country{United States}
  }

\author{Yiqiao Zhang}
\email{zhangyiqiao@pku.edu.cn}
\affiliation{%
  \institution{Peking Union Medical College Hospital}
   \city{Beijing}
  \state{}
  \country{China}
  }

\author{Qianyu Zhuang}
\email{zhuangqianyu@pumch.cn}
\affiliation{%
  \institution{Peking Union Medical College Hospital}
   \city{Beijing}
  \state{}
  \country{China}
  }

\renewcommand{\shortauthors}{Zhang, Li et al.}

\begin{abstract}
With the increasing integration of Multimodal Large Language Models (MLLMs) into the medical field, comprehensive evaluation of their performance in various medical domains becomes critical. However, existing benchmarks primarily assess general medical tasks, inadequately capturing performance in nuanced areas like the spine, which relies heavily on visual input. To address this, we introduce SpineBench, a comprehensive Visual Question Answering (VQA) benchmark designed for fine-grained analysis and evaluation of MLLMs in the spinal domain. SpineBench comprises 64,878 QA pairs from 40,263 spine images, covering 11 spinal diseases through two critical clinical tasks: spinal disease diagnosis and spinal lesion localization, both in multiple-choice format. SpineBench is built by integrating and standardizing image-label pairs from open-source spinal disease datasets, and samples challenging hard negative options for each VQA pair based on visual similarity (similar but not the same disease), simulating real-world challenging scenarios. We evaluate 12 leading MLLMs on SpineBench. The results reveal that these models exhibit poor performance in spinal tasks, highlighting limitations of current MLLM in the spine domain and guiding future improvements in spinal medicine applications. SpineBench is publicly available at https://zhangchenghanyu.github.io/SpineBench.github.io/.
\end{abstract}

\begin{CCSXML}
<ccs2012>
   <concept>
       <concept_id>10010147.10010178.10010224</concept_id>
       <concept_desc>Computing methodologies~Computer vision</concept_desc>
       <concept_significance>500</concept_significance>
       </concept>
   <concept>
       <concept_id>10010405.10010444.10010449</concept_id>
       <concept_desc>Applied computing~Health informatics</concept_desc>
       <concept_significance>500</concept_significance>
       </concept>
   <concept>
       <concept_id>10010405.10010444.10010447</concept_id>
       <concept_desc>Applied computing~Health care information systems</concept_desc>
       <concept_significance>300</concept_significance>
       </concept>
 </ccs2012>
\end{CCSXML}

\ccsdesc[500]{Computing methodologies~Computer vision}
\ccsdesc[500]{Applied computing~Health informatics}
\ccsdesc[300]{Applied computing~Health care information systems}

\keywords{Spinal Pathologies; Multimodal Large Language Models; Vision-Language Benchmark; Medical VQA}
\begin{teaserfigure}
  \centering
  \includegraphics[width=0.9\textwidth]{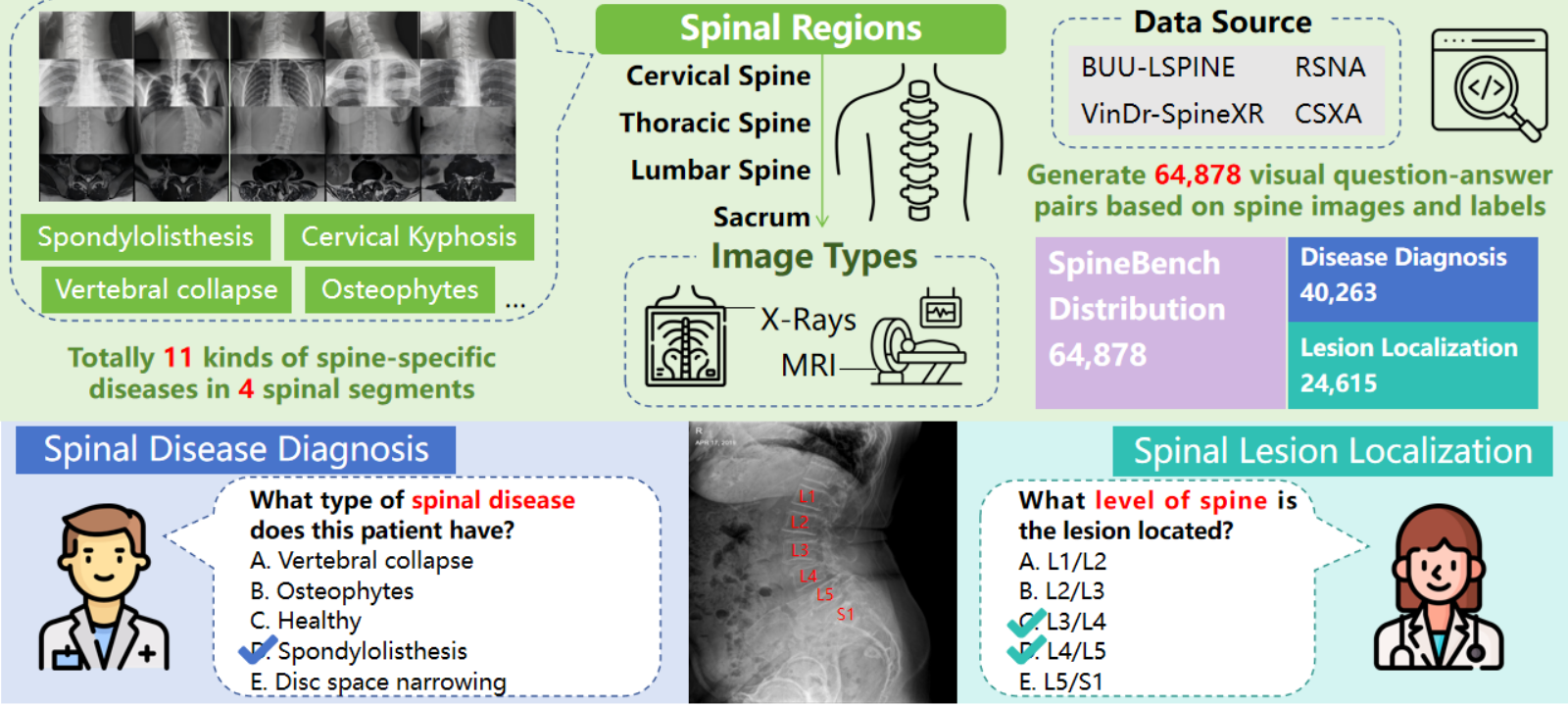}
  \caption{SpineBench includes 64,878 visual question-answer pairs from diverse data sources, covering 11 spine-specific diseases across 5 spinal segments, and introduces two tasks: Spinal Disease Diagnosis and Spinal Lesion Localization.}
  % {Overview of SpineBench. SpineBench comprises 64,878 visual question-answer pairs derived from multiple data sources covering 11 spine-specific diseases across 4 spinal segments and introduces two novel spine-specific vision analysis tasks: Spinal Disease Diagnosis and Spinal Lesion Localization.}
  \Description{SpineBench includes 64,878 visual question-answer pairs from diverse data sources, with the modality of X-ray and MRI. It covers 11 spine-specific diseases across 5 spinal segments. In SpineBench, we introduce two tasks: Spinal Disease Diagnosis and Spinal Lesion Localization. For the former task, we have 40,263 VQA pairs in it, while for the latter, we have 24,615 in the task.}
  \label{fig:teaser}
\end{teaserfigure}

%\received{31 May 2025}
%\received[revised]{31 July 2025}
%\received[accepted]{31 July 2025}

\maketitle

\section{Introduction}
\label{sec:intro}
Multimodal Large Language Models (MLLMs) have shown significant application potential within the medical domain~\cite{hartsock2024vision, qin2022medical}. 
However, existing medical benchmarks~\cite{yu2025medframeqa, ye2024gmai, xia2024cares, bao2024bmad, le2025u2} primarily assess MLLMs in general clinical scenarios, with limited focus on their performance within specialized medical fields.
% However, existing medical benchmarks~\cite{zuo2025medxpertqa, ye2024gmai, xie2024medtrinity} primarily focus on assessing the overall performance of MLLM in general medical contexts, often overlooking their efficacy within specific medical domains. 
Furthermore, the inherent disciplinary silos between medical sub-domains~\cite{albert2022barriers} significantly complicate the integration and utilization of domain-specific knowledge. Taking the spinal domain as an example, clinical practice heavily relies on a fine-grained semantic understanding of spinal images and precise segmentation and identification of vertebrae.
However, the absence of spine-specific evaluation datasets hinders the assessment of these models in spinal scenarios, limiting reliable evaluation of their real-world applicability~\cite{saeed2024msff, saravi2022artificial, tumko2024neural}.
Therefore, developing evaluation benchmarks specifically tailored for the spine domain is crucial to drive in-depth research and practical application of MLLMs in this field.

To address this gap, we introduce SpineBench, a large-scale, comprehensive Visual Question Answering (VQA) benchmark specifically designed for the spinal domain.
Given the inconsistency in annotation standards across existing spinal datasets, we first collect and systematically standardize image-label pairs from multiple sources to establish a unified and reliable foundation. Building upon this standardized dataset, we establish two tasks: \textbf{spinal disease diagnosis}, which classifies the disease types given the spine image (X-rays or MRI), and \textbf{spinal lesion localization}, which further identifies the segment and location within the spinal image that has lesions. This design deeply integrates real-world spinal clinical diagnostic workflows by evaluating not only what the spinal disease is but also providing a fine-grained assessment of where the disease occurs.

We format both tasks as multiple-choice questions. We employ Gemini-2.5-Pro~\cite{GoogleDeepMind2025Gemini} to generate diverse questions and sample similar spine images with different disease types to build hard negative distracting options that are different diseases but have similar visual appearances in spine images, simulating challenging real-world scenarios. In addition, while spinal disease diagnosis has only one correct answer, spinal lesion localization questions may have multiple correct options (lesion locations) to evaluate precise and comprehensive identification of spinal disease locations.

Finally, all generated samples undergo rigorous manual verification and calibration by domain experts to ensure clinical accuracy and linguistic quality. The resulting SpineBench contains 64,878 spine-specific VQA pairs, covering 11 diagnostic categories and lesion localization across five lumbar vertebrae. To further promote evaluation efficiency, we perform systematic quality assessment and carefully select 2,128 diverse and high-quality instances (1,000 for spinal disease diagnosis and 1,128 for spinal lesion localization), for fast yet comprehensive evaluation of MLLMs.

We evaluate 12 publicly available leading MLLMs, including 8 general-purpose MLLMs and 4 specialized medical MLLMs. Experimental results reveal that even advanced MLLMs have significant limitations on the SpineBench benchmark, highlighting substantial room for improvement in spine domain applications. Additionally, we further invite medical experts to evaluate the reasoning processes of models, focusing on  logical rigor and the application of professional expertise. This expert evaluation confirms deficiencies in both spine domain-specific knowledge and logical reasoning.

Overall, the contributions of this work include:
\begin{itemize}[left=0pt]
\item We propose a comprehensive and challenging benchmark for evaluating MLLMs in the spinal domain, featuring two specialized tasks (disease diagnosis and lesion localization) with multiple-choice questions that include highly distracting options and potentially multiple correct answers, simulating real-world diagnostic workflows and challenges.
\item We utilize a novel distractor sampling approach leveraging visual similarity to produce confusable options, better reflecting real-world diagnostic challenges where visually similar diseases pose identification difficulties.
\item Our evaluation of 12 prominent MLLMs reveals significantly limited performance, with most models achieving accuracy close to random guessing, indicating substantial limitations in the spinal domain. We employ human experts to evaluate model reasoning processes, further revealing deficiencies in both spine domain-specific knowledge and logical reasoning.
\end{itemize}

\section{Related Works}
\label{relate}

\subsection{Multimodal LLMs in medical domains}
Multimodal Large Language Models (MLLMs) like GPT~\cite{achiam2023gpt}, Gemini~\cite{GoogleDeepMind2025Gemini}, and numerous open-source models (e.g., InternVL~\cite{chen2024far, team2023internlm}, LLaVA~\cite{liu2023visual, liu2024llavanext}, Qwen~\cite{bai2023qwen, yang2024qwen2}) have shown significant medical potential. By training and fine-tuning Multimodal medical data, researchers have developed Medical MLLMs (Med-MLLMs) such as LLaVA-Med~\cite{li2023llava}, Med-Flamingo~\cite{moor2023med}, HealthGPT~\cite{lin2025healthgpt} and MedM-VL~\cite{shi2025medm}, extending general MLLMs to medicine. Med-MLLMs are also being applied to specific medical sub-fields; For example, EyecareGPT~\cite{li2025eyecaregpt} uses adaptive resolution to accommodate ophthalmic imaging. 
However, in spine medicine, MLLM applications face considerable deficiencies due to the scarcity of high-quality datasets and dedicated evaluation benchmarks.

% In recent years, Multimodal Large Language Models (MLLMs), represented by series such as GPT~\cite{achiam2023gpt}, DeepSeek\cite{lu2024deepseek, guo2025deepseek}, and numerous open-source models (e.g., InternVL~\cite{chen2024far, team2023internlm}, LLaVA~\cite{liu2023visual, liu2024llavanext}, Qwen~\cite{bai2023qwen, yang2024qwen2}), have demonstrated significant potential in the medical domain. By training and fine-tuning Multimodal data spanning multiple medical disciplines, researchers have developed Medical Multimodal Large Language Models (Med-MLLMs) like LLaVA-Med~\cite{li2023llava} and Med-Flamingo~\cite{moor2023med}, thereby extending the general MLLMs into medical domains. Besides, Med-MLLMs are gradually extending their applications into specific medical sub-fields. For example, EyecareGPT~\cite{li2025eyecaregpt} incorporates an adaptive resolution mechanism to accommodate variable image resolutions in clinical ophthalmic imaging. However, within the domain of spine medicine, the application of MLLM still faces considerable challenges due to the scarcity of high-quality datasets and dedicated evaluation benchmarks.

\subsection{Medical Benchmarks}
% Medical evaluation benchmarks are either domain-specific or general-purpose. Domain-specific benchmarks (e.g., VQA-RAD~\cite{lau2018dataset} and SLAKE~\cite{liu2021slake} for radiology, LMOD~\cite{qin2024lmod} for ophthalmology and PathVQA~\cite{he2020pathvqa} for pathology) focus on specific disciplines. In contrast, general-purpose medical benchmarks, such as GAMI-MMBench~\cite{ye2024gmai}, OmniMedVQA~\cite{hu2024omnimedvqa}, MMMU~\cite{yue2024mmmu} and MedFrameQA~\cite{yu2025medframeqa} assess capabilities across diverse medical domains. While current benchmarks cover general medical tasks and some specific disciplines, to the best of our knowledge, spine medicine lacks dedicated evaluation tasks for MLLMs' domain-specific knowledge.
Existing medical evaluation benchmarks are either domain-specific or general-purpose. Domain-specific benchmarks target particular medical specialties, including VQA-RAD~\cite{lau2018dataset} and SLAKE~\cite{liu2021slake} for radiology, LMOD~\cite{qin2024lmod} for ophthalmology, and PathVQA~\cite{he2020pathvqa} for pathology. General-purpose benchmarks such as GAMI-MMBench~\cite{ye2024gmai}, OmniMedVQA~\cite{hu2024omnimedvqa}, MMMU~\cite{yue2024mmmu}, and MedFrameQA~\cite{yu2025medframeqa} evaluate MLLM performance across multiple medical domains simultaneously.
Despite this broad coverage of medical specialties, spine medicine remains notably underrepresented, lacking dedicated evaluation benchmarks that can assess MLLMs' domain-specific knowledge and reasoning capabilities in spinal diagnostics.

% In the medical domain, existing evaluation benchmarks are categorized into two types: domain-specific benchmarks and general-purpose medical benchmarks. Domain-specific benchmarks focus on evaluating model capabilities within medical disciplines. For example, VQA-RAD~\cite{lau2018dataset} and SLAKE~\cite{liu2021slake} are utilized to evaluate MLLMs' understanding of radiological images; LMOD~\cite{qin2024lmod} is designed to test performance in specific ophthalmic applications; PathVQA~\cite{he2020pathvqa} concentrates on the field of pathology. In contrast, general-purpose medical benchmarks, such as GAMI-MMBench~\cite{ye2024gmai}, OmniMedVQA~\cite{hu2024omnimedvqa}, and MMMU~\cite{yue2024mmmu}, aim to achieve a more comprehensive assessment of models' capabilities across diverse medical domains. Although existing medical benchmarks have expanded to cover general medical tasks and evaluations for some specific medical disciplines, to the best of our knowledge, there remains a lack of evaluation tasks specifically targeting MLLMs' command of domain-specific knowledge in spine medicine.

\subsection{Spinal Multimodal Datasets}
High-quality spinal datasets are key to developing specialized diagnostic models. However, due to data barriers in the spine domain~\cite{ahmad2024datasets}, high-quality ones are often proprietary and not public. Most existing open-source spinal datasets~\cite{liebl2021computed, saeed2023automated, an2022part, chen2024vertxnet, li2023icunet++, van2024lumbar} are designed for spinal segmentation. Datasets with rich spinal disease information, exact diagnostic labels, and site annotations, are far rarer. Available spinal disease classification datasets~\cite{rsna-2024-lumbar-spine-degenerative-classification, klinwichit2023buu, natalia2022automated} often cover only 3-5 disease types, and their classification criteria may deviate from clinical practice. Therefore, a comprehensive, diverse, and meticulously annotated spinal disease dataset is urgently needed to support relevant MLLM evaluation.

\begin{figure}
\centering 
\includegraphics[width=0.9\columnwidth]{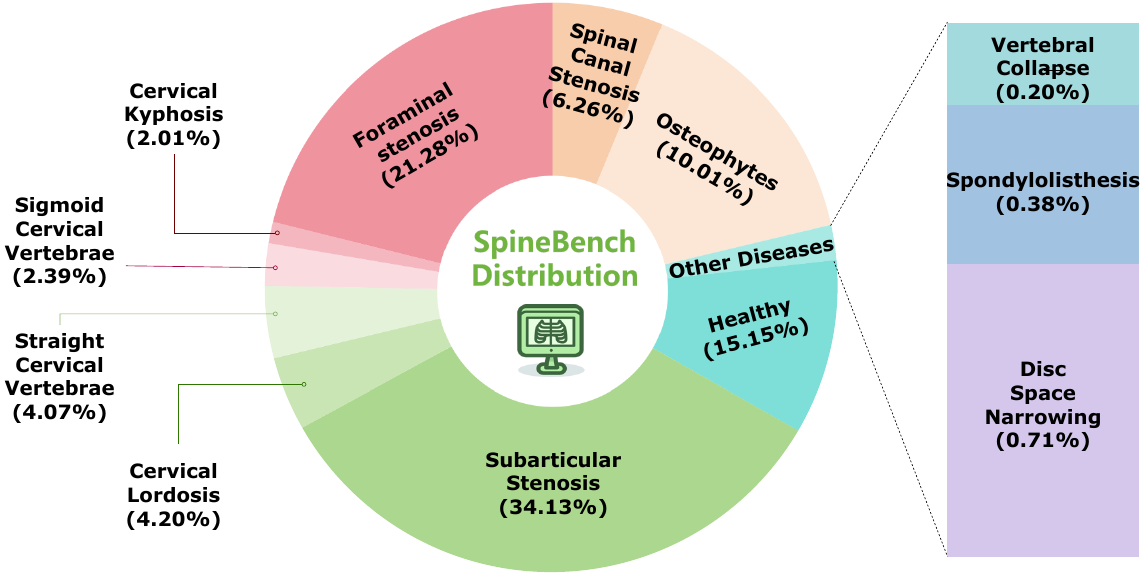}
% \vspace{-6mm}
\caption{Disease distribution in SpineBench. }
% Open-source samples for certain diseases are exceptionally scarce, resulting in the imbalance of disease distribution in SpineBench. } 
% \Description{...}
% \vspace{-6mm}
\label{img:data} 
\end{figure}

\begin{table}[ht!]
\centering
\caption{SpineBench Dataset Statistics}
\label{tab:dataset_stats}
\resizebox{0.475\textwidth}{!}{
\footnotesize
\begin{tabular}{lcc}
\toprule
\textbf{Component}  & \textbf{Evaluation} & \textbf{Total}\\
\midrule
\textbf{Total spine images} & \textbf{2,020} & \textbf{40,263}\\
Images with disease type annotations & 1,128 & 40,263\\
Images with lesion localization annotations & 1,000 & 24,615\\
\midrule
\textbf{Total VQA pairs} & \textbf{2,128} & \textbf{64,878}\\
VQA pairs for disease diagnosis  & 1,000 & 40,263\\
VQA pairs for lesion localization & 1,128 & 24,615\\
\midrule
\multicolumn{3}{l}{\textbf{Dataset Coverage}} \\
Disease types covered & \multicolumn{2}{c}{\textbf{11} (Distribution in Figure~\ref{img:data})} \\
Lumbar spinal segments & \multicolumn{2}{c}{\textbf{5} (L1/L2, L2/L3, L3/L4, L4/L5, L5/S1)} \\
\bottomrule
\end{tabular}
}
\end{table}

\section{SpineBench} 
\label{method}

\subsection{Data Collection and Standardization}
\begin{figure} 
\centering 
\includegraphics[width=0.9\columnwidth]{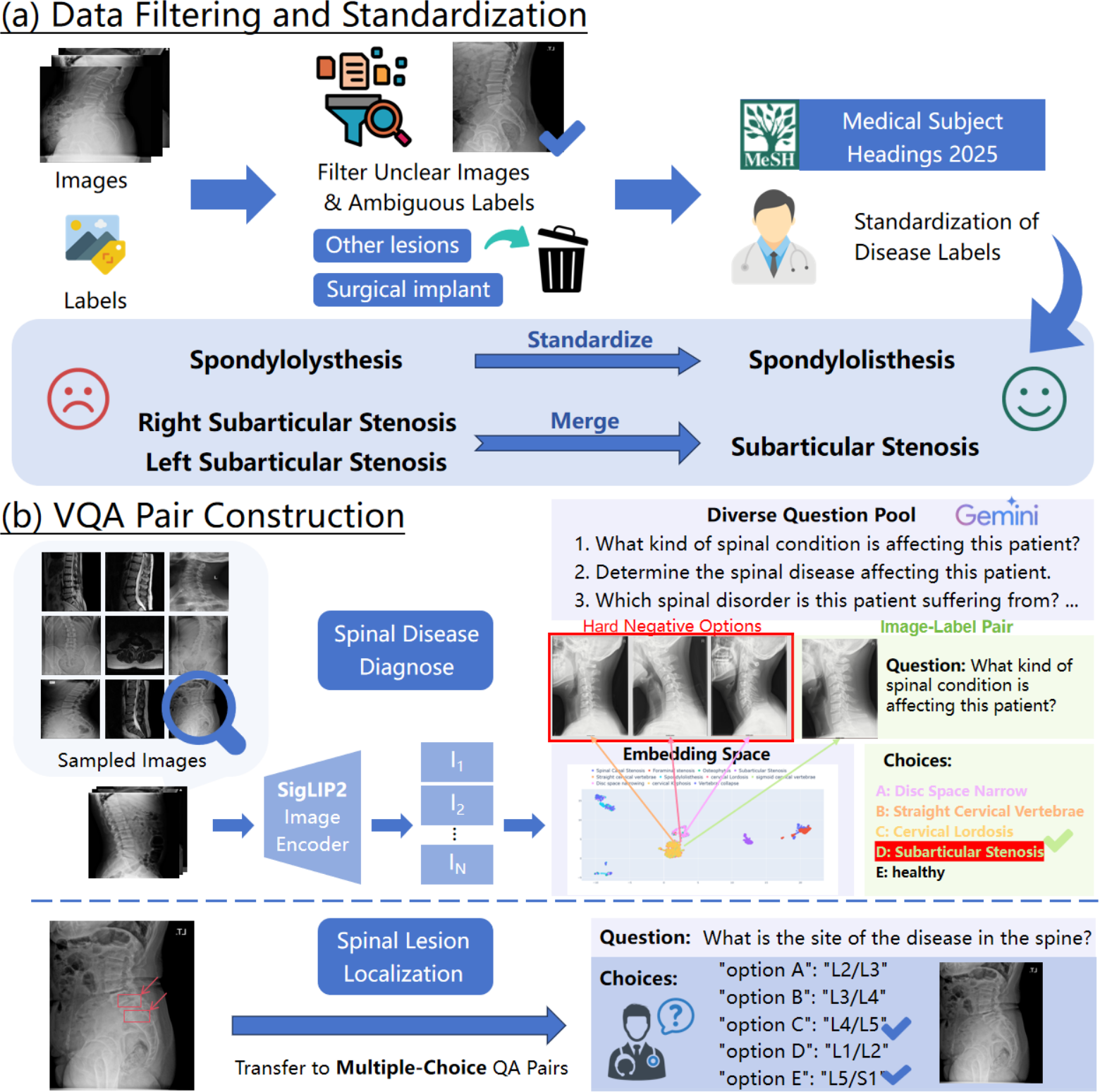}
% \vspace{-6mm}
\caption{SpineBench is constructed with two steps: benchmark construction and standardization (upper), and VQA pairs generation (lower).}
% The two key steps to construct SpineBench. The upper is the pipeline of Benchmark Construction and Standardization while the lower is the pipeline of VQA Pairs Generation.} 
\Description{...}
% \vspace{-6mm}
\label{liucheng1} 
\end{figure}

% As shown in Fig.\ref{liucheng1}. SpineBench is constructed in two key steps:
% \subsubsection{Data Collection and Standardization.}
Prior spinal research has primarily focused on spine segmentation, with public datasets for disease diagnosis and classification remaining scarce, as most are proprietary~\cite{ahmad2024datasets}. To address this gap, we integrate four spinal disease datasets encompassing both X-ray and MRI modalities: BUU Spine Dataset~\cite{klinwichit2023buu}, CSXA~\cite{ran2024high}, RSNA~\cite{rsna-2024-lumbar-spine-degenerative-classification}, and VinDr-SpineXR~\cite{nguyen2021vindr, pham2021vindr} from public resources to construct SpineBench.
SpineBench covers 11 distinct spinal diseases for diversity, and to maintain quality and consistency, we standardize both images and labels through a rigorous preprocessing pipeline. As illustrated in Figure~\ref{liucheng1}(a), at the image level, all 2D and 3D spinal images are converted to 2D RGB format. We meticulously remove images where the disease features are visually inconspicuous or whose labels lack clear clinical significance (e.g. ``Other lesions" or ``surgical implant"). At the label level, referencing the internationally authoritative Medical Subject Headings (MeSH) and the guidance of collaborating senior clinicians, we perform a detailed review, standardization, and revision of the disease labels from the original datasets. We also merge labels that distinguish anatomical orientation for the same disease (e.g., left-sided vs. right-sided lesion) for clarity and independence of disease definitions.

As shown in Table~\ref{tab:dataset_stats}, the resulting dataset comprises 40,263 spine images, each annotated with one of the 11 disease types. Among these, 24,615 images also have detailed lesion localization annotations across five lumbar spinal segments (L1/L2, L2/L3, L3/L4, L4/L5, and L5/S1). The distribution of the disease is shown in Fig.\ref{img:data}. Given limited images for some conditions, cervical-only diseases are grouped as ``Cervical Diseases," and those under 1\% of total images as ``Other Diseases". We then construct one VQA pair for each image in the dataset for the spinal disease diagnosis task, and a VQA pair for each image with location annotations for the spinal lesion localization task, using the approach described below.

\subsection{VQA Pair Construction}
Clinically, spinal disease diagnosis and lesion localization are coupled tasks. The spinal disease diagnosis task evaluates the model's capability to classify disease types given spine images (i.e., what is the disease), whereas the spinal lesion localization task focuses on spatial identification of affected regions (i.e., where the disease is located within the spine images).
Together, these two tasks provide a fine-grained evaluation of MLLMs' capability to identify spinal diseases from medical images.
We construct multiple-choice questions for both tasks, with the approach illustrated in Figure~\ref{liucheng1}(b).

%\textbf{Step 2: VQA Pairs Generation. }In clinical practice, the diagnosis of spinal disease and the localization of lesions are generally coupled processes. However, the Spinal Disease Diagnosis task aims to evaluate the model's analytical and reasoning abilities based on medical image, whereas the spinal lesion localization task focuses on evaluating its potential for image segmentation and spatial localization. To enable a fine-grained evaluation of MLLMs, we deconstruct these two aspects into distinct benchmark tasks and employ differentiate generation strategies in the subsequent construction of VQA pairs. The generation pipeline is illustrated in Figure~\ref{liucheng1}(b):

% \begin{itemize}[left=0pt]
\subsubsection{Diverse Question Generation.} The core query intents for the two tasks are straightforward (the former concerns ``What is the diagnosis?" and the latter addresses ``Where is the lesion located?"). However, people express the same intent using varied phrasings. To simulate such real-world query scenarios, we use Gemini-2.5-Pro to generate seven semantically equivalent questions with diverse expressions for both tasks. These questions form our question pool for VQA pair sampling. 

% \item \textbf{Diverse Question Formulation.} Although the core query intents for the two tasks are relatively straightforward (the former concerns "What is the diagnosis?" and the latter cares for "Where is the lesion located?"), in real-world interactions, individuals typically express the same intent using varied natural language phrasings. To evaluate the MLLMs' deep understanding of the semantic essence of questions, we aim for diversity in question formulation. For the core query intent of each task, we use Gemini-2.5-Pro to generate seven distinct questions that are semantically equivalent but vary in their expression. These generated questions form our question pool for the subsequent random sampling of VQA pairs.

\subsubsection{Image Similarity-based Distractor Option Sampling.} 
% \item \textbf{Image Similarity-based Distractor Option Generation.} To simulate the realistic clinical challenge where physicians must perform differential diagnosis among diseases with similar symptomatic presentations, we devise a distractor option generation pipeline based on image visual similarity. Specifically, we first randomly sample 200 representative images for each spinal disease from the whole dataset (if a disease has fewer than 200 samples, all are selected). Subsequently, SigLIP2~\cite{tschannen2025siglip} is used to extract feature vectors from these sampled images. These feature vectors are integrated and stored in a dedicated database along with the corresponding image metadata, thus constructing an embedding dataset. For each query image $I_q$ to be evaluated within the benchmark, its feature vector $v_q$ is extracted similarly. To identify the most visually similar disease categories as distractor options, we compute the cosine similarity between $v_q$ and the feature vectors $v_e$ of all images $i_e$ in the embedding dataset. Let $S(v_q,v_e)$ denote the cosine similarity, calculated as:

% For each disease category $C_k$ in the embedding dataset, the overall similarity $Sim(I_q,C_k)$ between the query image $I_q$ and this category $C_k$ is defined as the maximum cosine similarity between $I_q$ and any image $I_{e,j}$ within category $C_k$. And then select the top three disease categories $\{C_1^*, C_2^*, C_3^*\}$ with the highest similarity scores $Sim(I_q,C_k)$ as potential distractor options:

The challenge in real-world spinal disease diagnosis lies in the fact that different diseases may have similar visual appearances in spine images. To simulate such challenging scenarios, we propose an approach to sample hard negative options that have similar visual characteristics but different diagnoses.
Our dataset contains 11 distinct disease types. For each query image $I_q$ associated with a disease type $C_q$, we aim to sample three additional disease options plus the fixed option ``Healthy'' to construct a five-choice question, where only one option $C_q$ is correct.

First, we sample at most 200 images per spinal disease. Subsequently, we use SigLIP2~\cite{tschannen2025siglip} to extract feature vectors from these sampled images. The features of these sampled images constitute the embedding space for all disease types.
For each image $I_q$ used to build question-answer pairs for the spinal disease diagnosis task, its embedding $v_q$ is also extracted using SigLIP2. To identify visually similar disease categories as distractors, we compute the cosine similarity between $v_q$ and each embedding $v_e$ in the embedding dataset via cosine similarity, which denotes $S(v_q,v_e)$, calculated as:
\begin{equation}\label{eqn-5} 
  S(v_q,v_e) = \frac{\mathbf{v_q} \cdot \mathbf{v_e}}{|\mathbf{v_q}| |\mathbf{v_e}|}.
\end{equation}
% where $S(v_q,v_e)$ denote their similarity.

For each disease category $C_k$, we define the similarity $Sim(I_q, C_k)$ between the query image $I_q$ and category $C_k$, as the maximum cosine similarity between $I_q$ and any image $I_{e,j}$ within $C_k$ in the embedding space. We then select the top three disease categories $\{C_1^*, C_2^*, C_3^*\}$ with the highest $Sim(I_q, C_k)$ scores as distractor options:

% \begin{align}
%   Sim(I_q, C_k) &= \max_{I_{e,j} \in C_k} \{ S(\mathbf{v}_q, \mathbf{v}_{e,j}) \},  \\
%   \{C_1^*, C_2^*, C_3^*\} &= \underset{C_k, k \in \{1, \dots, N_{\text{disease}}\}}{\text{arg top-3}} \{( Sim(I_q, C_k) \}.   
% \end{align}

% \begin{align}
%   Sim(I_q, C_k) &= \max_{I_{e,j} \in C_k} S(\mathbf{v}_q, \mathbf{v}_{e,j}),  \\
%   \{C_1^*, C_2^*, C_3^*\} &= \arg\max_{C_k}^{(3)} Sim(I_q, C_k),
% \end{align}

\begin{align}
Sim(I_q, C_k) &= \max_{I_{e,j} \in C_k} S(\mathbf{v}_q, \mathbf{v}_{e,j}),  \\
\{C_1, C_2, C_3\} &= \underset{C_{k \in {1, \dots, 10}}}{\text{top-3}} Sim(I_q, C_k).
\end{align}

These three hard negative distractor disease types, together with the ground-truth disease type $C_q$ for the image $I_q$, and a ``Healthy'' option, form challenging and realistic five-choice questions for the spinal disease diagnosis task. "Healthy" is mandatorily included as a fixed option to evaluate between normal and abnormal discrimination, and the benchmark includes healthy spinal samples. This strategy generates highly confusable, visually plausible distractors, thereby effectively probing model diagnostic capabilities.
% build the challenge and realisitic five-choice questions for the spinal disease diagnosie task.
% After excluding the ground-truth disease label of $I_q$,these visually similar categories become distractor options for VQA pairs, forming the Answer Pool. "Healthy" is mandatorily included as a fixed option to evaluate between normal and abnormal discrimination, and the benchmark includes healthy spinal samples. This strategy generates highly confusable, visually plausible distractors, thereby effectively probing model diagnostic capabilities.

% After excluding the ground-truth disease label of the query image $I_q$, these visually similar disease categories are selected as distractor options for the VQA pair, collectively constituting the Answer Pool for each QA pair. Furthermore, to evaluate the ability of MLLMs to discriminate between normal and abnormal conditions, "Healthy" is mandatorily included as a fixed option in the candidate answers for every VQA pair, while ensuring the benchmark contains a certain number of healthy spinal samples. This strategy aims to generate highly confusable, visually plausible distractors, thereby more effectively probing and challenging the models' true diagnostic capabilities.
    
% \end{itemize}

\subsubsection{Multi-Label Lesion Localization Task.}
Regarding the spinal lesion localization task, spinal lesions can span multiple vertebral segments among the five locations (L1/L2, L2/L3, L3/L4, L4/L5, L5/S1). This means that the spinal lesion localization task may have multiple correct answers. We construct VQA pairs with the five fixed options corresponding to these spinal segments. Different images have their ground-truth lesion annotations, which form the correct answer set that could contain one or multiple segments.

To prevent the model from learning positional patterns, we randomly shuffle the option order in each prompt.
% This ensures that models must rely on content understanding rather than positional biases.
Ultimately, we construct 64,878 high-quality VQA pairs, comprising 40,263 for the spinal disease diagnosis task and 24,615 for the spinal lesion localization task.

\subsection{Data Split}
\begin{figure}
\centering 
\includegraphics[width=0.9\columnwidth]{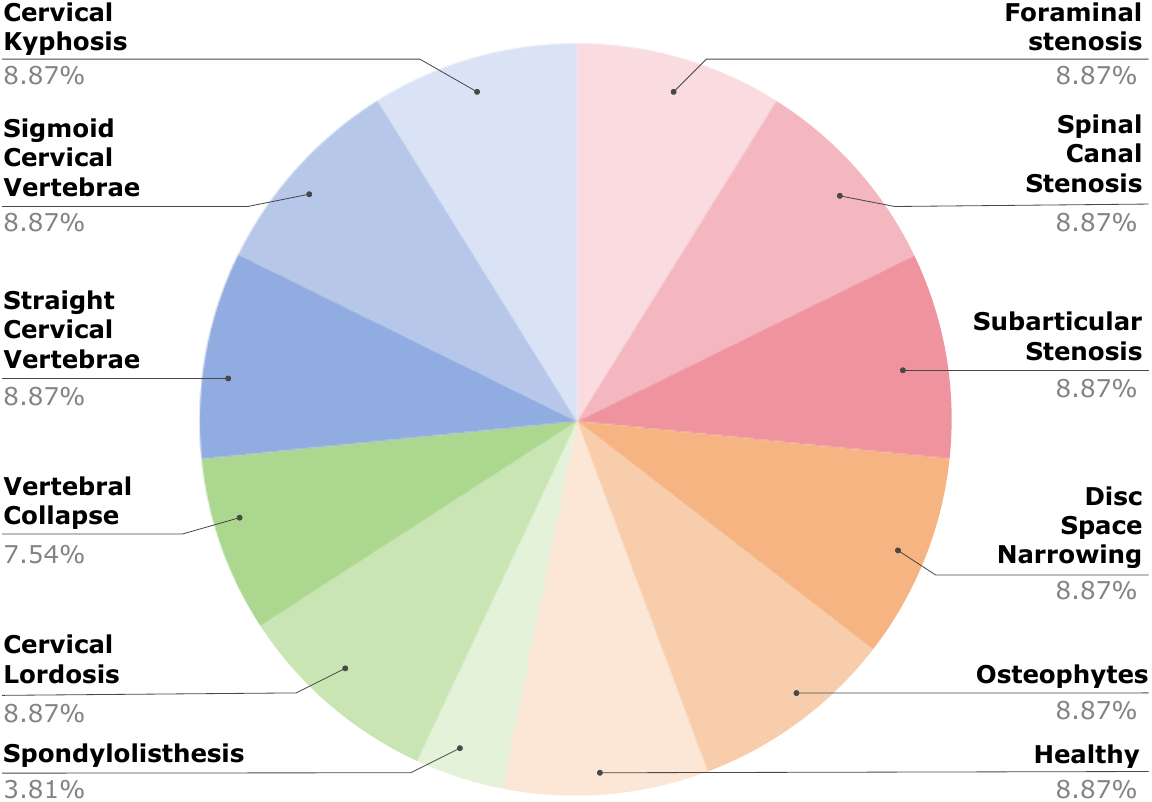}
\caption{Disease distribution of the evaluation set.}
\label{img:data2} 
\end{figure}

 To enhance evaluation efficiency and fairness, a subset is sampled from SpineBench using a meticulous filtering pipeline designed to ensure both visual quality and class balance.
 First, we assess the image quality for each data entry based on objective metrics such as variance, Laplacian variance, and contrast. Entries that satisfy predefined quality standards are then grouped according to their primary disease category. 
 Following this, we employ an iterative selection strategy that prioritizes the selection of candidate data entries for disease categories that currently exhibit insufficient sample sizes. 
 Subsequently, these candidate entries are evaluated through a specially designed scoring mechanism, This mechanism favors entries whose potentially incorrect options corresponded to disease categories that have, up to that point, appeared less frequently as distractors. This measure is intended to promote not only a balanced distribution of target diseases but also robust diversity and balance among the disease categories involved in its distractor options. Representative images are then selected from this pool for each disease category, striving to balance both the number of images per disease and the distribution of distractors. 
 Furthermore, to ensure data quality, we perform an additional manual verification to ensure the validity of the benchmark. The final subset comprises 1,000 cases for the spinal disease diagnosis task and 1,128 cases for the spinal lesion localization task, whose detailed data distribution is presented in Figure ~\ref{img:data2}. 
 
\subsection{Evaluation Metrics}
For the spinal disease diagnosis task, which has only one correct answer, we use the standard accuracy ($\text{Acc}$) as the evaluation metric. Let $n_{\text{correct}}$ denote the number of questions for which the model provided the correct answer, and $N$ denotes the total number of questions. The accuracy is calculated as: 
% For the spinal disease diagnosis task, which has only one correct answer, we use the standard accuracy ($\text{Acc}$) as the evaluation metric. Let $n_{\text{correct}}$ denote the number of questions for which the model provided the correct answer, and $N$ denote the total number of questions. The accuracy is calculated as:
\begin{equation}\label{eqn-1} 
  \text{Acc} = \frac{n_{\text{correct}}}{N} 
\end{equation}

For the spinal lesion localization task, which involves multiple correct answers corresponding to lesion locations, we employ instance-based evaluation metrics: accuracy ($\text{Acc}$), precision ($\text{Pre}$), and recall ($\text{Rec}$) to assess both the accuracy and comprehensiveness of the predictions. 
For each question $i$, we define: $n_{\text{match}}^{(i)}$ as the number of correctly predicted options that match the ground truth, $l_{\text{pred}}^{(i)}$ as the total number of options predicted by the model, and $l_{\text{gt}}^{(i)}$ as the total number of ground truth options. The instance-based metrics are calculated as:
\begin{align}
    \text{Acc} &= \frac{1}{N} \sum_{i=1}^{N} \mathbf{1}(n_{\text{match}}^{(i)} = l_{\text{gt}}^{(i)} \text{ and } l_{\text{pred}}^{(i)} = l_{\text{gt}}^{(i)}) \label{eqn-2}\\
    \text{Pre} &= \frac{1}{N} \sum_{i=1}^{N} \frac{n_{\text{match}}^{(i)}}{l_{\text{pred}}^{(i)}}, \quad \text{Recall} = \frac{1}{N} \sum_{i=1}^{N} \frac{n_{\text{match}}^{(i)}}{l_{\text{gt}}^{(i)}}\label{eqn-3}
\end{align}
where $N$ is the total number of questions, and $\mathbf{1}(\cdot)$ is the indicator function that returns 1 if the condition is true and 0 otherwise.
A comprehensive description of the evaluation protocol is provided in Appendix.

\section{Experiments}
\label{experiment}
\subsection{Experimental Details}
We evaluate eight general-purpose MLLMs: Gemini-2.5-pro-exp-03-25~\cite{GoogleDeepMind2025Gemini}, Gpt-4o-2024-11-20~\cite{hurst2024gpt}, Qwen2.5-VL-7B-Instruct~\cite{bai2025qwen2}, Claude-3-5-sonnet-20241022~\cite{anthropic2024computer}, Llama-3.2-11b-vision-instruct~\cite{Meta2024Llama}, Grok 3~\cite{xAI2025Grok}, Doubao-1.5-vision-pro-32k~\cite{ByteDance2025Doubao} and Deepseek-VL2~\cite{wu2024deepseek}; and four Med-MLLMs: LLaVA-Med v1.5~\cite{li2023llava}, HuatuoGPT-Vision~\cite{chen2024huatuogpt}, Med-Flamingo~\cite{moor2023med} and MedGemma~\cite{medgemma-hf}. Experimental environments and hyperparameter settings followed the official source code for each model. To ensure optimal model performance across different evaluation tasks, we design custom prompts for the two core tasks: spinal disease diagnosis and spinal lesion localization. We employ zero-shot prompting and encourage the model to provide reasoning before delivering final answers.

\subsection{Main Experiments}
% The detailed results of MLLM evaluation on SpineBench are summarized in Table~\ref{tab:main_results}. Overall, the experimental findings clearly indicate that SpineBench poses a 
% significant challenge to current MLLMs. 
% % Even when models generate plausible reasoning, their diagnostic and localization accuracy remains highly limited.
% For spinal disease diagnosis, excluding the best-performing model, Gemini-2.5-Pro-Exp-03-25, which achieves only 32.37\% accuracy, most MLLMs achieves accuracy close to random guessing. 
% The performance on lesion localization is similarly poor.
% % , with most models struggling to accurately identify affected spinal segments.
% % On spinal lesion localization, MLLM performance is even weaker. 
% Although these models achieve respectable precision and recall, most models struggle to match all ground-truth options, leading to low standard accuracy. Figure~\ref{fig:point} highlights this, showing a gap between the precision and recall values for most models. This means models often identify some affected vertebrae may be guess but fail to identify all. Overall, the accuracy on two tasks are extrenely low, even thought the best model acihive just 20.86\%. These results highlight significant room for improvement in MLLMs' ability to comprehend complex medical images, particularly in identifying and localizing subtle spinal lesions.
% The performance on lesion localization is similarly poor, with most models struggling to accurately identify affected spinal segments.

\begin{figure} 
\centering 
\includegraphics[width=0.9\columnwidth]{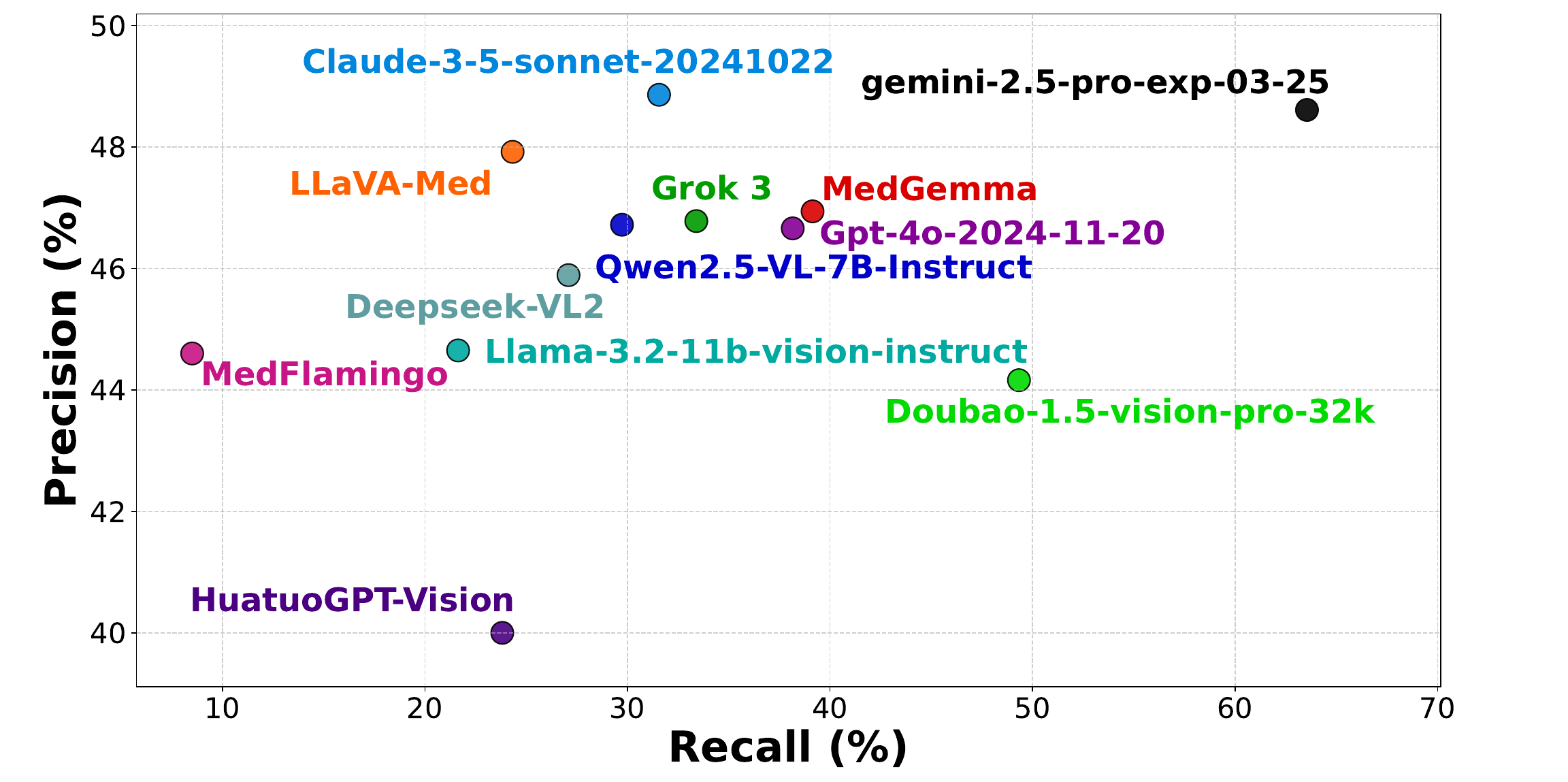}
% \vspace{-6mm}
\caption{Model performance on Spinal Lesion Localization.} 
\Description{Description of the image.}
% \vspace{-6mm}
\label{fig:point} 
\end{figure}

The detailed results of MLLM evaluation on SpineBench are summarized in Table~\ref{tab:main_results}. Overall, the experimental findings clearly indicate that SpineBench poses a significant challenge to current MLLMs. Even if models might generate plausible reasoning, their diagnostic and localization accuracy is highly limited. For spinal disease diagnosis, excluding the best-performing model, Gemini-2.5-Pro-Exp-03-25, which achieves only 32.37\% accuracy, most MLLMs achieve accuracy close to random guessing.
The performance on lesion localization is similarly poor. Although these models achieve reasonable precision and recall scores, most models struggle to match all ground-truth options, leading to low exact-match accuracy. Figure~\ref{fig:point} highlights this phenomenon, showing a substantial gap between precision/recall values and accuracy for most models. This indicates that models often identify some affected vertebrae correctly but fail to identify the complete set of lesions. Overall, the accuracy on both tasks is extremely low, with even the best model achieving just 20.86\% accuracy.
These results highlight significant room for improvement in MLLMs' ability to comprehend complex medical images, particularly in identifying and localizing subtle spinal lesions.

% The detailed results of the MLLM evaluation on SpineBench benchmark are summarized in Table~\ref{tab:main_results}. Overall, the experimental findings clearly indicate that SpineBench poses a formidable challenge to current state-of-the-art MLLMs. Even if some models could generate seemingly plausible reasoning processes, their final diagnostic and localization accuracy remained highly limited. In the spinal disease diagnosis task, the best performing model, gemini-2.5-pro-exp-03-25, achieved an accuracy of only 32.37\%. The diagnostic accuracy of most MLLMs fails to significantly surpass the level of random guessing, and some models potentially perform even below this baseline. This highlights substantial room for improvement in the ability of MLLMs to comprehend complex medical images, especially in identifying and localizing subtle lesions. 

\begin{table}
    \centering
%     \tiny
    \caption{Model performance on SpineBench. \textbf{SDD} denotes Spinal Disease Diagnose. In each column, the best and second-best performance are marked in \textcolor{red}{red} and \textcolor{blue}{blue}, respectively.}
    % The best-performing model in each category is \textcolor{red}{red}, and the second best is \textcolor{blue}{blue}.}
    \label{tab:main_results}
    % \vspace{-3mm}
    \resizebox{0.95\linewidth}{!}
    {
    \begin{tabular}{lccccc}
        \toprule
        \multirow{2}{*}{\textbf{Models}} & \textbf{SDD} & \multicolumn{3}{c}{\textbf{Spinal Lesion Localization}} & \textbf{Overall}\\ \cmidrule(lr{0pt}){2-2} \cmidrule(lr{0pt}){3-5} \cmidrule(lr{0pt}){6-6}
         & Acc & Acc& Pre & Recall & Acc \\
        \midrule
        Random & 20.00\% & - & - & - & - \\
        \rowcolor{orange!20}
        \multicolumn{6}{c}{\textit{Generalist Models}} \\
        Gemini-2.5-pro-exp-03-25~\cite{GoogleDeepMind2025Gemini} & \textcolor{red}{32.37\%} & 9.35\% & \textcolor{blue}{48.61\%} & \textcolor{red}{63.55\%} & \textcolor{red}{20.86\%} \\ 
        Gpt-4o-2024-11-20~\cite{hurst2024gpt} & \textcolor{blue}{23.64\%} & 9.69\% & 46.66\% & 38.16\% & 16.67\% \\ 
        Claude-3-5-sonnet-20241022~\cite{anthropic2024computer} & 23.58\% & \textcolor{blue}{12.64\%} & \textcolor{red}{48.86\%} & 31.56\% & \textcolor{blue}{18.11\%} \\ 
        Llama-3.2-11b-vision-instruct~\cite{Meta2024Llama} & 19.24\% & 10.00\% & 44.65\% & 21.64\% & 14.62\% \\ 
        Grok 3~\cite{xAI2025Grok} & 21.45\% & 10.02\% & 46.78\% & 33.40\% & 15.74\% \\ 
        Doubao-1.5-vision-pro-32k~\cite{ByteDance2025Doubao} & 17.46\% & 6.14\% & 44.16\% & \textcolor{blue}{49.33\%} & 11.80\% \\ 
        DeepSeek-VL2~\cite{wu2024deepseek} & 23.25\% & 11.39\% & 45.89\% & 27.09\% & 17.32\% \\ 
        Qwen2.5-VL-7B-Instruct~\cite{bai2025qwen2} & 18.26\% & 12.14\% & 46.72\% & 29.73\% & 15.20\% \\
        \rowcolor{gray!20}
        \multicolumn{6}{c}{\textit{Medical Models}}  \\
        LLaVA-Med v1.5~\cite{li2023llava} & 21.32\% & 9.25\% & 47.92\% & 24.33\% & 15.29\% \\ 
        HuatuoGPT-Vision-7B~\cite{chen2024huatuogpt} & 22.28\% & 8.80\% & 40.00\% & 23.82\% & 15.84\% \\ 
        Med-Flamingo~\cite{moor2023med} & 19.49\% & 4.08\% & 44.60\% & 8.51\% & 11.79\% \\ 
        MedGemma-4B~\cite{medgemma-hf} & 12.15\% & 5.30\% & 46.94\% & 39.14\% & 8.73\%  \\ 
        \bottomrule
    \end{tabular}
    }
    % \vspace{-3mm}
\end{table}

% The cause may be insufficient complex multi-select reasoning samples in training data or limited advanced cognitive abilities (like multi-step reasoning, evidence integration, exclusion principles), highlighting an urgent need for targeted optimization.

% In the spinal lesion localization task, the performance of existing Large Vision-Language Models (MLLMs) is even more markedly deficient. Although these models achieve respectable micro-averaged accuracy and recall, which suggests a rudimentary capability to identify lesion sites and comprehend basic spatial relationships, most MLLMs still struggle to correctly match all ground-truth options, leading to generally low standard accuracy on such tasks. Focusing specifically on micro-averaged metrics, as illustrated in Figure~\ref{fig:point}, a discernible gap exists between the $Recall_{mcq}$ and $Acc_{mcq}$ values for most models. This indicates that while models can often identify some of the vertebrae affected by disease, they generally fail to identify all of them comprehensively. This broader phenomenon profoundly reveals deficiencies in current models when processing complex logical relationships among options (e.g., mutual exclusivity, inclusion, dependency). The root cause may lie in the insufficient representation of such complex, multi-select reasoning samples in training data, or limitations in the models' own advanced cognitive abilities, such as multi-step reasoning, evidence integration, and the application of exclusion principles, highlighting the urgent need for targeted optimization.

\subsection{MLLM Reasoning Evaluation}

To assess the reasoning capabilities of MLLMs, we invite collaborating clinical physicians to evaluate model reasoning processes in spinal disease diagnosis and lesion localization tasks across four dimensions: 
\textbf{1) Clinical Plausibility}: This dimension evaluates whether the model's reasoning aligns with clinical logic and remains consistent with established medical knowledge, diagnostic principles, and anatomical understanding relevant to spinal pathology. 
\textbf{2) Identification and Utilization of Key Visual Features}: This assesses whether the reasoning accurately identifies and prioritizes relevant visual evidence in the image to support diagnostic conclusions. 
\textbf{3) Depth of Pathological Understanding}: This determines whether the reasoning reflects superficial pattern matching or demonstrates deeper comprehension of the specific pathophysiology presented. 
\textbf{4) Quality of Spatial Reasoning} (spinal lesion localization task only): This evaluates the reasoning process's ability to demonstrate an understanding of 3D spatial relationships within the spine. 
LLaVA-Med and Med-Flamingo are excluded from this evaluation due to their excessively brief reasoning outputs. Detailed scoring rubrics are provided in the Appendix. Figure~\ref{fig:human} presents the clinicians' average scores on a 0-5 scale for the various MLLMs.

\begin{figure} 
\centering 
\includegraphics[width=0.9\columnwidth]{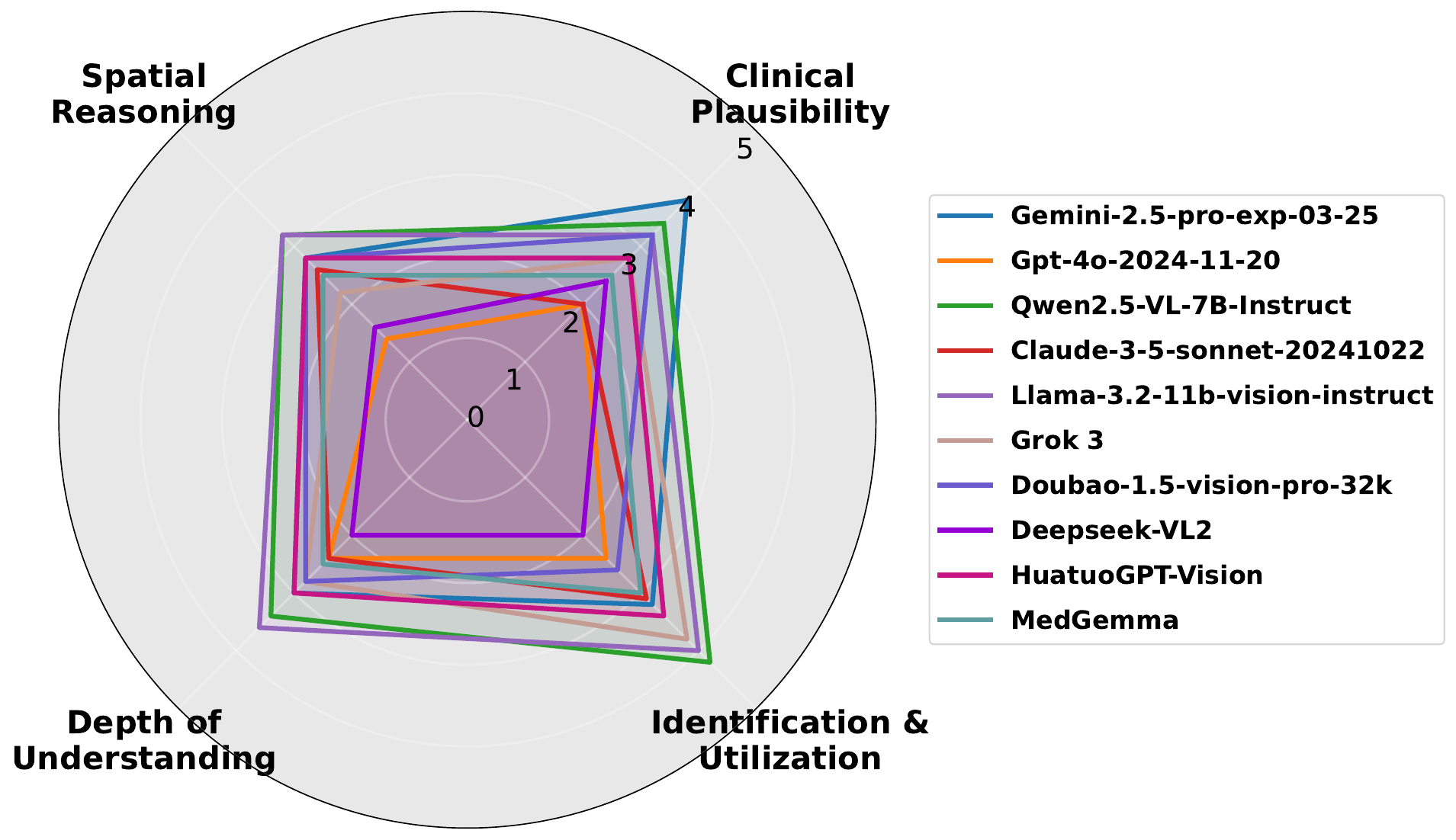}
% \vspace{-6mm}
\caption{MLLM Reasoning Average Scores (Max score: 5) from medical experts. Most models scored below 4, indicating limited capability in spatial pathological reasoning.} 
\Description{Description of the image.}
% \vspace{-3mm}
\label{fig:human} 
\end{figure}

The evaluation results reveal significant differentiation among MLLMs across the four reasoning dimensions. While most MLLMs demonstrate basic clinical reasoning plausibility and understand spinal 3D spatial relationships, their understanding of specific spinal pathologies remains superficial. Most models can identify approximate lesion locations but fail to provide deep interpretation or explain the underlying pathophysiological mechanisms.

% The evaluation results revealed significant differentiation among MLLMs of varying performance levels. Although most MLLMs have established a certain foundational level of clinical reasoning plausibility and can understand the three-dimensional spatial relationships of the spine, their understanding of specific spinal pathologies is generally not thorough. The majority of models can only superficially indicate the approximate location of lesions, failing to provide a deeper interpretation and explanation of their pathophysiological mechanisms.

% \subsubsection{Insufficient Fine-grained Understanding of Spinal Imagery.}
\subsection{Error Analysis}
\begin{figure} 
\centering 
\includegraphics[width=0.9\columnwidth]{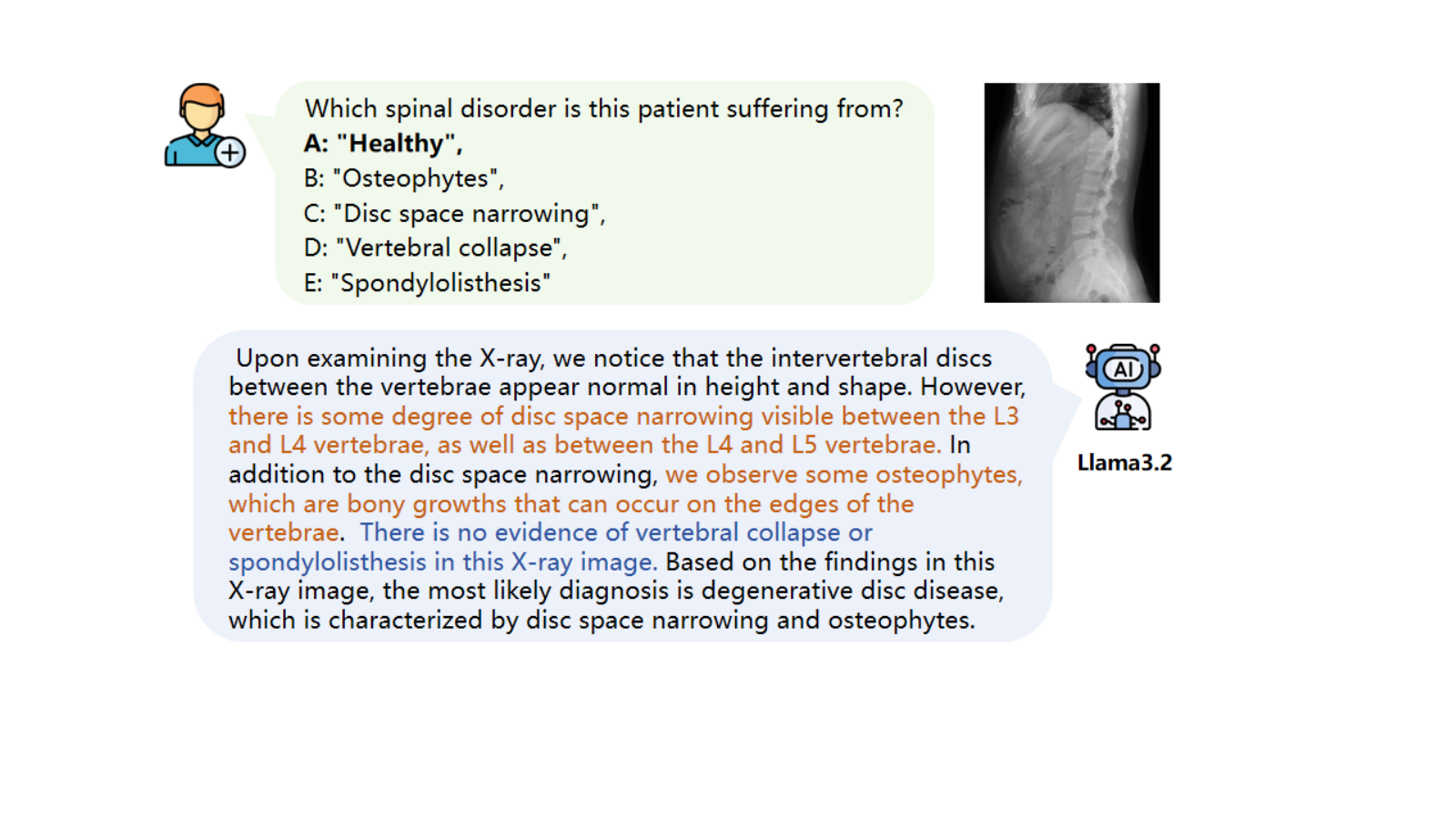}
% \vspace{-6mm}
\caption{Llama3.2 diagnose analysis case. \textcolor{brown}{brown} text indicates a wrong interpretation of image features, and \textcolor{blue}{blue} indicates a correct judgment.} 
\Description{case}
% \vspace{-3mm}
\label{fig:case-in} 
\end{figure}

To further investigate performance bottlenecks and error causes, we analyze the reasoning processes and outputs of MLLM. Our findings reveal that current MLLMs have significant limitations in nuanced comprehension of spinal imagery. Specifically, most models demonstrate difficulty in identifying key pathological features within the images. For example, Figure~\ref{fig:case-in} illustrates an instance where an MLLM incorrectly interprets the normal lumbar curvature as pathological.

To test whether inadequate disease definition understanding contributes to poor performance, we augmented prompts with detailed definitions of all 11 diseases and re-evaluated the models. Results in Table~\ref{tab:new-evaluation} show limited improvement in diagnostic accuracy, indicating that the primary bottleneck lies in insufficient fine-grained understanding of spinal imagery rather than inadequate knowledge of disease definitions.
These findings suggest that current MLLMs require substantial improvement in visual semantic understanding of spinal pathology, which appears to be the key factor limiting diagnostic performance.

% \begin{table}
%     \caption{The evaluation of the Spinal Disease Diagnose with the definition of 11 spinal diseases into prompt.} % Optional: if you want a caption
%     \vspace{-3mm}
%     \label{tab:new-evaluation} % Optional: if you want to reference this table
%     \resizebox{\columnwidth}{!}{
%     \begin{tabular}{lccc}
%         \toprule
%         \textbf{Model} & \textbf{w/o Definition}  & \textbf{w/ Definition} & \Delta \\
%         \midrule
%         \rowcolor{orange!20}
%         \multicolumn{4}{c}{\textit{Generalist Models}} \\
%         Gemini-2.5-pro-exp-03-25~\cite{GoogleDeepMind2025Gemini} & 32.37\% & 29.24\%\\
%         Gpt-4o-2024-11-20~\cite{hurst2024gpt} & 23.64\% & 23.81\%\\ 
%         Qwen2.5-VL-7B-Instruct~\cite{bai2025qwen2} & 18.26\% & 15.96\% \\
%         Claude-3-5-sonnet-20241022~\cite{anthropic2024computer} & 23.58\% & 23.92\%  \\ 
%         Llama-3.2-11b-vision-instruct~\cite{Meta2024Llama} & 19.24\% & 22.37\% \\ 
%         Grok 3~\cite{xAI2025Grok} & 21.45\% & 20.61\% \\ 
%         Doubao-1.5-vision-pro-32k~\cite{ByteDance2025Doubao} & 17.46\% & 15.08\%  \\ 
%         DeepSeek-VL2~\cite{wu2024deepseek} & 23.25\% & 20.43\% \\ 
%         \rowcolor{gray!20}
%         \multicolumn{4}{c}{\textit{Medical Models}}  \\
%         LLaVA-Med v1.5~\cite{li2023llava} & 21.32\% & 17.13\% \\ 
%         HuatuoGPT-Vision-7B~\cite{chen2024huatuogpt} & 22.28\% & 21.97\% \\ 
%         Med-Flamingo~\cite{moor2023med} & 19.49\% & 18.48\% \\
%         MedGemma~\cite{medgemma-hf} & 12.15\% &  12.15\% \\
%         \bottomrule
%     \end{tabular}
%     }
%     \vspace{-6mm}
% \end{table}

\begin{table}
    \caption{Ablation study on the effect of including disease definitions in prompts for the Spinal Disease Diagnosis task.} % Optional: if you want a caption
    % \vspace{-3mm}
    \label{tab:new-evaluation} % Optional: if you want to reference this table
    \resizebox{0.95\columnwidth}{!}{
    \begin{tabular}{lccc}
        \toprule
        \textbf{Model} & \textbf{w/o Definition}  & \textbf{w/ Definition} & \textbf{$\Delta$} \\
        \midrule
        \rowcolor{orange!20}
        \multicolumn{4}{c}{\textit{Generalist Models}} \\
        Gemini-2.5-pro-exp-03-25~\cite{GoogleDeepMind2025Gemini} & 32.37\% & 29.24\% & -3.13\%\\
        Gpt-4o-2024-11-20~\cite{hurst2024gpt} & 23.64\% & 23.81\% & +0.17\%\\ 
        Claude-3-5-sonnet-20241022~\cite{anthropic2024computer} & 23.58\% & 23.92\% & +0.34\%\\ 
        Llama-3.2-11b-vision-instruct~\cite{Meta2024Llama} & 19.24\% & 22.37\% & +3.13\%\\ 
        Grok 3~\cite{xAI2025Grok} & 21.45\% & 20.61\% & -0.84\%\\ 
        Doubao-1.5-vision-pro-32k~\cite{ByteDance2025Doubao} & 17.46\% & 15.08\% & -2.38\%\\ 
        DeepSeek-VL2~\cite{wu2024deepseek} & 23.25\% & 20.43\% & -2.82\%\\ 
        Qwen2.5-VL-7B-Instruct~\cite{bai2025qwen2} & 18.26\% & 15.96\% & -2.30\%\\
        \rowcolor{gray!20}
        \multicolumn{4}{c}{\textit{Medical Models}}  \\
        LLaVA-Med v1.5~\cite{li2023llava} & 21.32\% & 17.13\% & -4.19\%\\ 
        HuatuoGPT-Vision-7B~\cite{chen2024huatuogpt} & 22.28\% & 21.97\% & -0.31\%\\ 
        Med-Flamingo~\cite{moor2023med} & 19.49\% & 18.48\% & -1.01\%\\
        MedGemma~\cite{medgemma-hf} & 12.15\% &  12.15\% & 0.00\%\\
        \bottomrule
    \end{tabular}
    }
    % \vspace{-5mm}
\end{table}

\section{Conclusion}
\label{con}
% We introduce SpineBench, a novel VQA benchmark to evaluate LVLMs' diagnostic reasoning and anatomical understanding in the spinal domain. SpineBench includes 64,878 QA pairs for spinal disease diagnosis and lesion localization, addressing a gap in evaluating LVLM applications in spine medicine. It innovatively uses image similarity for incorrect QA options, simulating clinical differential diagnosis and increasing the evaluative challenge. We evaluated 12 representative LVLMs. Results show that while models demonstrate some understanding and localization capabilities, significant performance bottlenecks still persist, especially in precise lesion localization. We believe SpineBench will drive development of next-generation, spine-specific LVLMs, enhancing their practical application and value in spine medicine.

This work introduces SpineBench, a comprehensive and challenging benchmark designed to evaluate and advance MLLMs in the spinal domain. The benchmark comprises 64,878 multiple-choice visual questions covering 11 spine-specific diseases and 5 spinal segments across two essential clinical tasks: spinal disease diagnosis and lesion localization, enabling fine-grained evaluation of model capabilities.
SpineBench presents unique challenges that closely simulate real-world diagnostic scenarios. The benchmark features carefully designed hard negative distractors based on visual similarity, while the lesion localization task supports multi-label answers to reflect the clinical reality where multiple spinal segments may be simultaneously affected. 
Our comprehensive evaluation of 12 leading MLLMs, including both general-purpose and medical-specific models, reveals significant limitations in current models' ability to comprehend complex spinal imagery. 
% Even advanced models achieve only modest performance, with the best model reaching just 32.37\% accuracy on disease diagnosis and 20.86\% on lesion localization. Human expert evaluation further confirms that while models demonstrate basic clinical reasoning, they lack deep pathological understanding and struggle with fine-grained visual feature identification.
% These findings indicate that the primary bottleneck lies in insufficient visual semantic understanding of spinal pathology rather than textual comprehension deficits. 
We believe SpineBench could serve as both a rigorous evaluation framework for assessing current capabilities and a valuable resource for enhancing future model development in specialized medical domains requiring nuanced visual interpretation.

\begin{acks}
This research is sponsored by National Natural Science Foundation of China (Grant No. 62306041), Beijing Nova Program (Grant No. 20230484488), National High Level Hospital Clinical Research Funding(2022-PUMCH-B-002), Beijing Natural Science Foundation(L222012). This work specially thanks our advisor, Terry Jianguo Zhang, for his invaluable guidance and support throughout this research.
\end{acks}

\bibliographystyle{ACM-Reference-Format}
\balance
\bibliography{sample-base}

\end{document}